\documentclass[conference]{IEEEtran}
\IEEEoverridecommandlockouts
\usepackage{cite}
\usepackage{amsmath,amssymb,amsfonts}
\usepackage{algorithmic}
\usepackage{booktabs}
\usepackage{enumitem}
\usepackage{graphicx}
\usepackage{framed,multirow}
\usepackage{float} 
\usepackage{subfigure} 
\usepackage{textcomp}
\usepackage{xcolor}
\newcommand{\ie}{\textit{i.e.}}
\newcommand{\eg}{\textit{e.g.}}
\newcommand{\etc}{\textit{etc}}
\newcommand{\etal}{\textit{et al.}}

\def\BibTeX{{\rm B\kern-.05em{\sc i\kern-.025em b}\kern-.08em
    T\kern-.1667em\lower.7ex\hbox{E}\kern-.125emX}}
\begin{document}

\title{Semi-Supervised Active Learning for COVID-19 Lung Ultrasound Multi-symptom Classification\\
}

\author{
\IEEEauthorblockN{1\textsuperscript{st} Lei Liu, 1\textsuperscript{st} Wentao Lei, \\ 3\textsuperscript{th} Xiang Wan, 4\textsuperscript{th} Li Liu$^{\ast}$\thanks{* Corresponding author}}
\IEEEauthorblockA{\textit{Shenzhen Research Institute of Big Data,} \\
\textit{The Chinese University of Hong Kong, Shenzhen}\\
Shenzhen, China \\
\{leiliu,wentaolei\}@link.cuhk.edu.cn, \\  wanxiang@sribd.cn, liuli@cuhk.edu.cn}
\and
\IEEEauthorblockN{1\textsuperscript{st} Yongfang Luo, 2\textsuperscript{nd} Cheng Feng}
\IEEEauthorblockA{\textit{Department of Medical Ultrasonics, }\\
\textit{National Clinical Research Center for Infectious Disease,} \\
\textit{Shenzhen Third People's Hospital (Second Hospital Affiliated}\\ \textit{ to Southern University of Science and Technology)}\\
Shenzhen, China \\
luoyongfang2005@foxmail.com, chaosheng-01@szsy.sustech.edu.cn}
}
\maketitle

\begin{abstract}
Ultrasound (US) is a non-invasive yet effective medical diagnostic imaging technique for the COVID-19 global pandemic. However, due to complex feature behaviors and expensive annotations of US images, it is difficult to apply Artificial Intelligence (AI) assisting approaches for the lung's multi-symptom (multi-label) classification. To overcome these difficulties, we propose a novel semi-supervised Two-Stream Active Learning (TSAL) method to model complicated features and reduce labeling costs in an iterative manner. The core component of TSAL is the multi-label learning mechanism, in which label correlation information is used to design a multi-label margin (MLM) strategy and a confidence validation for automatically selecting informative samples and confident labels. In this framework, a multi-symptom multi-label (MSML) classification network is proposed to learn discriminative features of lung symptoms, and a human-machine interaction (HMI) is exploited to confirm the final annotations that are used to fine-tune MSML. Moreover, a novel lung US dataset named COVID19-LUSMS is built, currently containing 71 clinical patients with 6,836 images sampled from 678 videos. Experimental evaluations show that TSAL can achieve superior performance to the baseline and the state-of-the-art using only 20\% data. Qualitatively, visualization of the attention map confirms a good consistency between the model prediction and the clinical knowledge.

\end{abstract}


\section{Introduction}
The novel coronavirus (COVID-19) has spread worldwide and is now officially a global pandemic. Typical diagnosing tools mainly include Computed tomography (CT) and X-ray, which are characterized by their relatively accurate performances \cite{wang2020clinical}. However, due to the prevalence of COVID-19, in practice, deep learning-based CT or X-ray approaches remain several challenges. Firstly, CT and X-ray tools are generally inflexible and involve extra radiations. Secondly, images of CT and X-ray are not easy to collect from COVID-19 patients because the imaging procedures involve isolated patients, complex clinical equipment, and many other nontrivial processes. 

In contrast, lung ultrasound (US) imaging is preferred as a mature tool for its fast, flexible, and reliable deployment, especially in emergencies \cite{bourcier2016lung}. More importantly, it is non-invasive and can work at the bedside. Recently, some works \cite{soldati2020proposal,huang2020a,Qian2020Findings} focused on COVID-19 symptom detection based on lung US images. Indeed, based on lung US images, automatic AI assisting approaches for COVID-19 symptoms classification are significant for medical diagnoses. Therefore, we focus on lung US multi-symptom classification in this work.

\begin{figure}[!t]
    \setlength{\abovecaptionskip}{0.cm}
	\centering
	\includegraphics[width=1\linewidth]{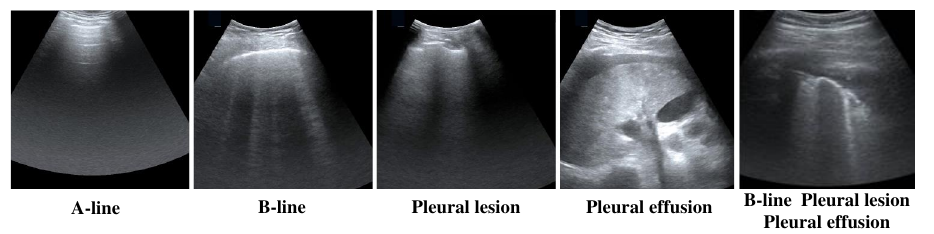}
    \caption{Examples of the COVID19-LUSMS dataset.}
	\label{example}
	\vspace{-2.0em}
\end{figure}

In practice, the automatic classification of COVID-19 lung symptoms is difficult for twofold reasons. Firstly, the lung US images of COVID-19 patients may simultaneously present multiple symptoms, which exhibit complicated image features (see Fig. \ref{example}). One possible solution is the multi-label learning, which targets to judge whether an image possesses multiple characteristics denoted by labels \cite{Liang2018Attentive,guo2017human}. 
Secondly, it is expensive and tedious to collect and annotate numerous COVID-19 lung US images. To address this difficulty, a feasible solution is active learning \cite{liu2017active,vezhnevets2012active}, which aims to achieve satisfactory performance given a limited labeling cost.

\begin{figure}[!t]
    \setlength{\abovecaptionskip}{0.cm}
	\centering
	\includegraphics[width=0.95\linewidth]{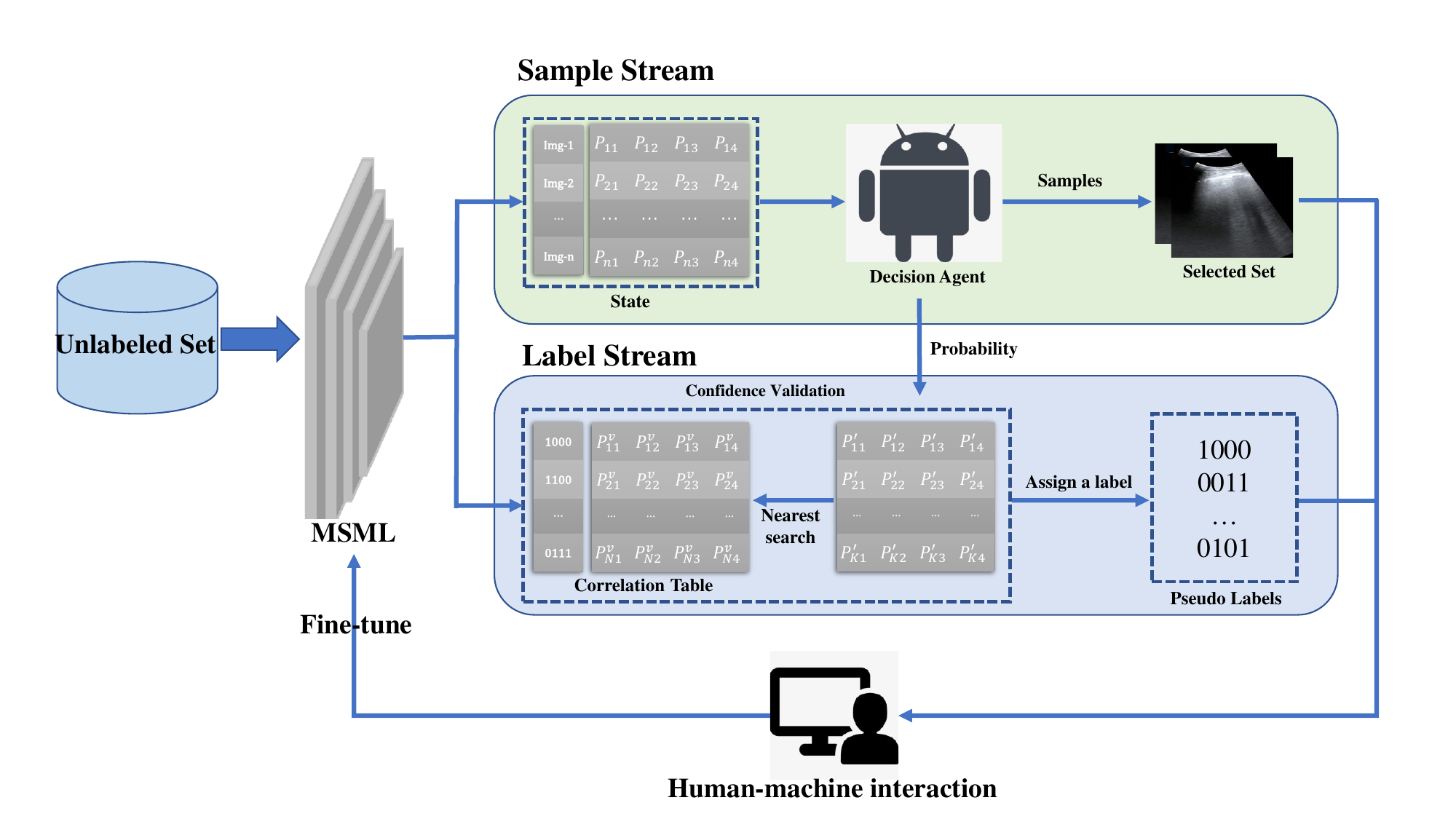}
	\caption{Overview of the proposed MSML-TSAL. Sample stream is to select informative samples by state and decision agent, which are used to describe the prediction results of all unlabeled images and determine which image candidate will be selected. Label stream is to validate and assign pseudo labels for the selected images. Human-machine interaction is to confirm the final annotations for model updating. This iterative operation terminates when it reaches the annotation budget or required performance.}
	\label{framework}
	\vspace{-1.0em}
\end{figure}

In this paper, we propose a novel semi-supervised Two-Stream Active Learning (TSAL) method, which works iteratively by sample selection, pseudo-label validation, human-machine interaction and CNN parameters updating. In TSAL, a multi-symptom multi-label (MSML) classification network is constructed as the basic model for feature learning. The sample stream works for informative sample selection by newly designed multi-label margin strategy (MLM), while the label stream is exploited to assign confident pseudo-label for selected images. Then HMI is used for confirming the final annotations to fine-tune the MSML. An overview of the proposed method is shown in Fig. \ref{framework}.

Besides, a large-scale dataset of lung US images for COVID-19 is built for this work. Some examples are shown in Fig. \ref{example}. Experiments on this dataset show that our proposed method achieves superior classification performance, outperforming the baseline models and the state-of-the-art (SOTA) using less than 20\% of the labeled images. 
To the best of our knowledge, this is the first work on automatic multi-symptom multi-label classification for COVID-19 lung US images. 

In summary, this work contains the following three contributions:
\begin{enumerate}
     \item A novel large-scale dataset containing lung US images of COVID-19 is built specifically for this work. This dataset is annotated in the multi-label form by medical experts.
    \item We propose a novel semi-supervised TSAL method, which effectively reduces the labeling cost. It exploits label correlation information to select informative samples and confident pseudo labels.
    \item Experimental results show that our method achieves superior performance using less than 20\% labeled data, compared with baselines and SOTA. Explainable analyses using attention map confirm that our results are well consistent with clinic expertise.
\end{enumerate}

\section{Related Work}
\subsection{Lung US techniques for COVID-19}
Many studies \cite{Bourcier2014Performance,bourcier2016lung,Claes2017Performance} reported the superiority of US imaging in diagnosing pneumonia and related lung conditions. Sloun \etal \cite{8805336} proposed a fully convolutional neural network (CNN) \cite{he2016deep,Simonyan15} to identify and localize the B-lines in clinical lung ultrasonography. 
Born \etal \cite{soldati2020proposal} trained a POCOVID-Net on a 3-class dataset and achieved good accuracy in classification. \cite{9093068} presented a novel deep network that simultaneously predicted the disease severity score associated with an input frame. However, they generally require a large-scale labeled US-image dataset, where annotations are expensive.

\subsection{Multi-label image classification}
As an important branch of classification, multi-label classification \cite{Tsoumakas2009Multi,zhang2014a} has been widely explored in recent years, supported by CNNs. It plays an important part in bridging the gap between low-level features and high-level semantic information \cite{Liang2018Attentive}. Some multi-label classification methods applied CNNs to obtain competitive performances. \cite{gong2013deep} optimized top-k ranking objectives combined with convolutional architectures to learn a better feature representation. Hypotheses-CNN-Pooling \cite{wei2014Cnn} incorporated object segmentation hypotheses with max pooling to generate multi-label predictions. For lung US images, a COVID-19 patient may perform multiple symptoms simultaneously. Thus, COVID-19 symptoms classification can be formulated as a multi-symptom multi-label classification task.

\subsection{Active learning}
AL has been successfully deployed into semantic segmentation \cite{sun2015active}, image classification \cite{beluch2018power}, human pose estimation \cite{liu2017active}, \etc. These applications indicate that AL is a great choice for labeling efforts reduction. Besides, there have been many popular selection strategies in the literature, mainly including query by committee \cite{10.1145/130385.130417}, expected error reduction \cite{article}, expected model change \cite{inproceedings}, and uncertainty sampling \cite{LEWIS1994148}. These strategies are usually exploited for single-label classification tasks, which are not suitable for COVID-19 lung US image classification with complex multi-label feature behaviors. In this work, we introduce AL into multi-symptom multi-label classification to actively reduce the labeling efforts. 

\begin{figure}[!t]
    \setlength{\abovecaptionskip}{0.cm}
	\centering
	\includegraphics[width=1\linewidth]{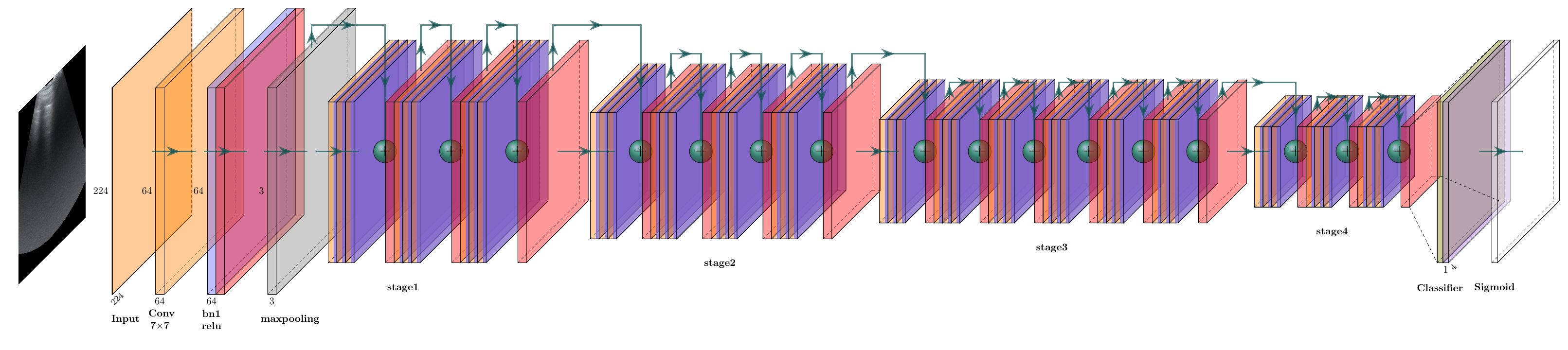}
    \caption{Architecture of the MSML model.}
	\label{structure}
	\vspace{-1.0em}
\end{figure}

\section{Methodology}
\subsection{Multi-symptom multi-label model}
The detailed architecture of the proposed MSML network is shown in Fig. \ref{structure}. The model backbone is ResNet50 \cite{he2016deep} pre-trained on ImageNet for its competitive performance and its relatively concise structure. Some modifications are made based on it, firstly, the hidden fully-connected layer is removed, and layers after the final convolutional layer are also removed to avoid too deep or too complicated architecture. Then it is followed by one specifically designed classifier consisting of average pooling fully-connected layer for multi-label tasks. To effectively learn the discriminative features of COVID-19 pulmonary symptoms, we adopt sigmoid cross-entropy loss, which can be calculated as:

\begin{equation}
 {\rm L_{CE}} = -\frac{1}{N}\sum\limits_{i=1}^{N}(y^{(i)}\log \hat{y}^{(i)} + (1-y^{(i)})\log (1-\hat{y}^{(i)})),
\label{cross}
\end{equation}
where $\hat{y}^{(i)}=1/1+e^{-x}$, and $y^{(i)}$ is the ground truth of the input, $N$ is the batch size and $x$ is the output of the last layer.

\subsection{Two-stream deep active learning framework}
The detailed framework of MSML-TSAL is presented in Fig. \ref{framework}. Firstly, we regard all unlabeled images as the candidate pool. At each AL iteration $t$, in the sample stream, the MSML network provides a prediction state $S_t$ for unlabeled images. Then decision agent makes an action for sample selection according to the state and selection strategy. Then pseudo labels are assigned by confidence validation in the label stream. The final annotations would be confirmed by HMI to fine-tune the MSML model. This iterative operation repeats until the expected performance of MSML or the empty candidate pool. 

\subsubsection{Sample stream}

\textbf{State}: The state is utilized to describe the relationship between unlabeled images and the prediction capability of the model. Prediction probability has been widely exploited to measure the prediction capability of the model in AL tasks. In this work, we exploit output prediction probabilities to construct the state matrix. At each AL iteration $t$, the candidate pool is denoted by $D=\{d_1, d_2, ... , d_{n}\}$, where $d_i$ is the $i$th unlabeled sample, and $n$ is the pool size. The prediction probability vectors are extracted by MSML model. The $i$th prediction vector can be written as $p_i = (p_{i1}, p_{i2}, ... , p_{il})^\mathrm{T}$, where $l$ is the number of the labels. The state matrix $S_t$ can be denoted as $S_t=(p_{1}, p_{2}, ... , p_{n})$.

\textbf{Action}
The action is to select unlabeled images from the candidate pool. At each iteration $t$, the selected set $A_t = \left\{ a_1, a_2, ..., a_{K_{max}} \right\}$ is decided by the decision agent according to state $S_t$, where $a_i$ is the $i$th selected sample, $K_{max}$ is the pre-defined annotation efforts (\ie, account of the labeled samples) for each AL iteration. Once an action is executed, the selected samples are removed from the candidate pool.

\textbf{Decision Agent}: The decision agent is used to measure which image is worth annotating using selection strategies. Firstly, we note that there are few active strategies for the multi-label classification task. To adapt to this task, we redesign two classical strategies of multi-class classification, including Least Confidence\cite{Settles09activelearning} and Entropy \cite{4563068}.

Least confidence (LC): Lower confidence of an image illustrates that it is hard for the classifier to make a correct prediction. The calculation of LC is:
\begin{equation}
 {\rm LC}(x) ={ \max \limits_{1\leq i \leq l}p(l_i\mid x)},
\end{equation}
where $l$ is the number of labels (\ie, symptoms). $p(l_i|x)$ denotes the prediction probability of symptom $l_i$ appearing in image $x$.

Multi-label entropy (MLE): Higher entropy indicates that the image carries rich information. The MLE is denoted as:
\begin{equation}
{\rm  MLE}(x) =\sum_{i=1}^l (p(l_i\mid x)\log p(l_i\mid x)+p(\overline{l_i}\mid x)\log p(\overline{l_i}\mid x)),
\end{equation}
where $p(l_i|x)$ is the probability of symptom $l_i$ appearing in image $x$ and $p(\overline{l_i}|x)$ = $1-p(l_i|x)$.

We find that LC prones to select the noisy samples and MLE doesn't consider the relation among different symptoms. Thus, we specially design a multi-label learning strategy called multi-label margin (MLM) to evaluate the informativeness of each unlabeled sample. The MLM is defined as:
\vspace{-1.0em}

\begin{equation}
{\rm MLM}(x) =\left| p(l_1\mid x)-\max \limits_{2\leq i \leq l}p(l_i\mid x)\right|,
\end{equation}
where $p(l_1|x)$ is the probability of A-line appearing in image $x$. For lung US images of COVID-19 patients, A-line denotes the health while others denote the disease. Thus, from the view of the medical knowledge, it is not reasonable that A-line appears with other symptoms simultaneously in an image. From the perspective of the model prediction, it is difficult to judge whether this image is healthy or unhealthy if the probability of A-line has a small margin with other symptoms, which may indicate that model has not learned effective information of this image. Thus, this margin intuitively can measure the informativeness of the unlabeled images.


\begin{figure}[!t]
    \setlength{\abovecaptionskip}{0.2cm}
	\centering
	\includegraphics[width=1\linewidth]{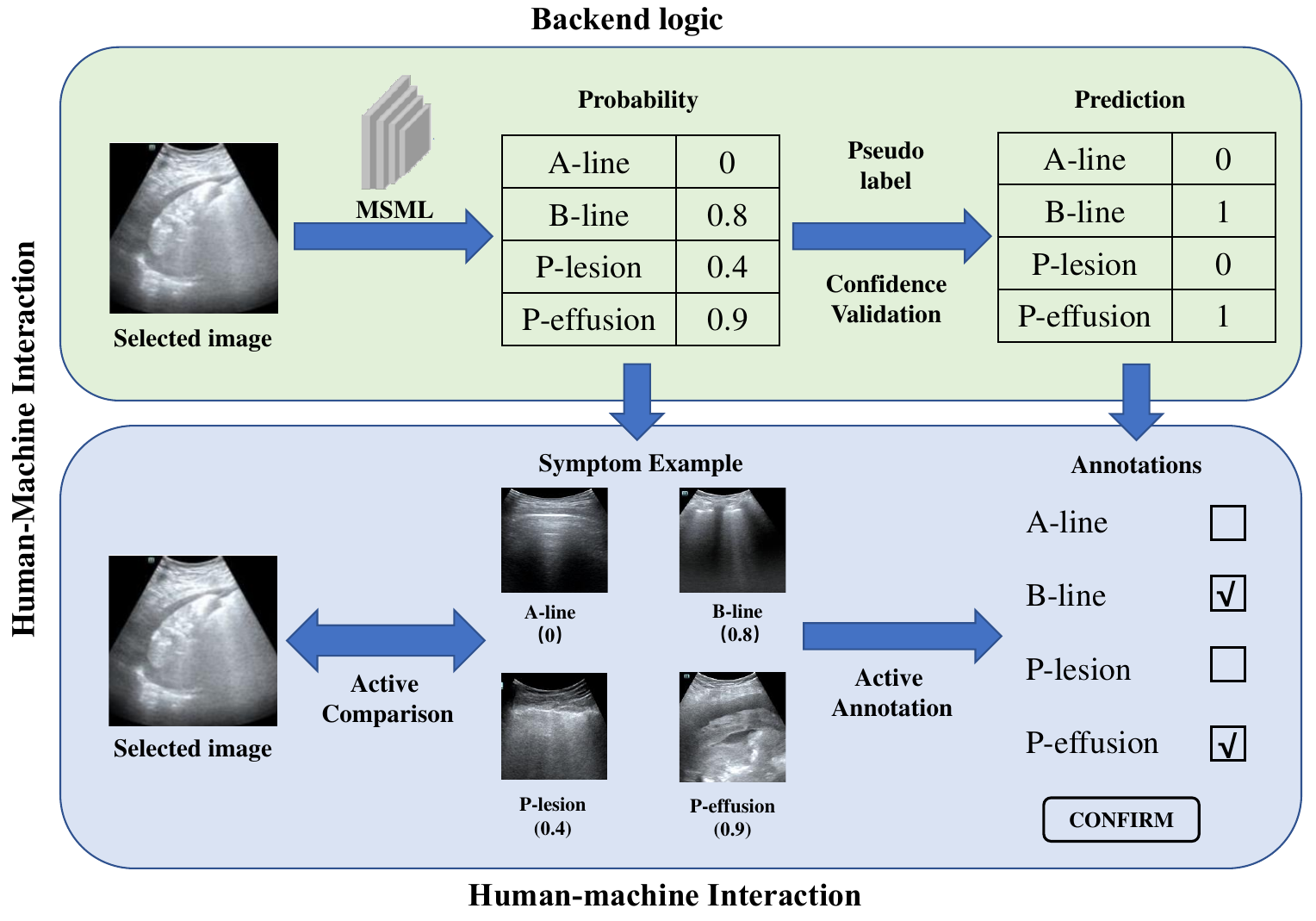}
    \caption{Overview of the human-machine interaction.}
	\label{interface}
	\vspace{-1.0em}
\end{figure}

\subsubsection{Label stream}
\textbf{Confidence validation}: Label correlations information has been widely employed for multi-label learning \cite{6471714} by mining the potential relationships among different labels, which also can be exploited for confidence validation. For convenience, we call ``0" or ``1" of a single symptom as a single label, and call ``0001" (``0110", ``0111", \etc) of four symptoms as label combination. In this work, we construct a correlation table to store the probability distribution information for each label combination. 

Given the selected and labeled samples at each AL iteration $t$, several probability matrices $\{P_1, P_2, \cdots, P_N\}$ can be built according to different label combinations. $N$ is the amount of the label combinations, which is $2^{l-1} + 1$. The matrix of $n$th label combination can be written as:
\begin{equation}
 P_n=
\begin{pmatrix}
    p_{11}&p_{12}&\cdots&p_{1l} \\
    p_{21} & p_{22} & \cdots & p_{2l} \\
    \vdots & \vdots & \vdots & \vdots \\
    p_{m1} & p_{m2} & \cdots & p_{ml} \\
\end{pmatrix}, \ n \in \{1, 2, \dots, N\},
\end{equation}
where $l$ is the number of the labels, $m$ is the number of the labeled samples with $n$th label combination.

The average probability vector $p^{avg}_{n}$ for $n$th probability matrix $P_n$ is calculated as:

\begin{table*}[!t]
	\renewcommand\arraystretch{1.1}
	\fontsize{8}{9}\selectfont
	\footnotesize
	\centering
	\caption{Comparisons with the baselines and SOTA. The bolds are of our method.}
	\begin{tabular}{cccccccccccccc}
		\toprule
	\multirow{2}{*}{Method} & \multicolumn{3}{c}{A-line} &\multicolumn{3}{c}{B-line} &\multicolumn{3}{c}{P-lesion}  &\multicolumn{3}{c}{P-effusion} &\multirow{2}{*}{data}                                                \\ \cmidrule(l){2-4} \cmidrule(l){5-7} \cmidrule(l){8-10} \cmidrule(l){11-13}
		& Acc & Sen & Spe  & Acc & Sen & Spe & Acc & Sen & Spe & Acc & Sen & Spe \\ 
		\cmidrule(l){1-1} \cmidrule(l){2-4} \cmidrule(l){5-7} \cmidrule(l){8-10} \cmidrule(l){11-13}
		VGG16     & 100 & 100 & 100 & 88.39 & 98.34 & 43.08 & 60.82 & 76.77 & 48.68 & 88.53 & 0 & 98.96 & 100\% \\
		
		ResNet34  & 100 & 100 & 100 & 90.88 & 96.87 & 63.63 & 73.36 & 81.71 & 67.00 & 89.31 & 0 & 99.84 & 100\%  \\
		ResNet50  & 100 & 100 & 100 & 88.60 & 98.95 & 41.50 & 80.34 & 82.86 & 78.41 & 89.45 & 0 & 100 & 100\%  \\ 
		ResNet101 & 100 & 100 & 100 & 90.45 & 98.17 & 55.33 & 79.91 & 82.37 & 78.04 & 89.45  & 0 & 100 & 100\% \\ \hline
		POCOVID-Net     & 100 & 100 & 100 & 84.97 & 90.35 & 60.47 & 80.84 & 79.90 & 81.55 & 91.02 & 14.86 & 100  & 100\% \\
		NNBD     & 100 & 100 & 100 & 90.31 & 99.91 & 46.64 & 71.86 & 68.20 & 74.65 & 89.45 & 0 & 100  & 100\% \\\hline
		MSML+TSAL(Random)  & 99.85 & 100 & 100 & 90.74 & 98.52 & 65.61 & 83.47 & 77.92 & 98.87 & 89.38 & 0 & 99.92 & 16.6\%  \\
	MSML+TSAL(MLE)  & 100 & 100 & 100 & 89.52 & 97.56 & 52.96 & 80.34 & 75.94 & 83.68 & 89.45 & 0 & 100 & 27.6\%  \\ 
		MSML+TSAL(LC) & 100& 100 & 100 & 94.30 & 95.91 & 86.95 & 83.19 & 78.74 & 86.57 & 89.45 & 0 & 100 & 16.6\% \\\hline
	\textbf{MSML}     & \textbf{100} & \textbf{100} & \textbf{100} & \textbf{95.72} & \textbf{98.78} & \textbf{81.81} & \textbf{80.98} & \textbf{81.38} & \textbf{80.67} & \textbf{90.09} & \textbf{6.08} & \textbf{100} & \textbf{100}\%\\
	\textbf{	MSML+TSAL(MLM)} & \textbf{100} & \textbf{92.38} & \textbf{100} & \textbf{98.50} & \textbf{98.79} & \textbf{92.49} & \textbf{83.26} & \textbf{76.77} & \textbf{96.36} & \textbf{89.45}  & \textbf{0} & \textbf{100} & \textbf{14.7\%}\\\hline
	\end{tabular}
	\label{table:com}
\end{table*}

\begin{equation}
p^{avg}_{n} = (\frac{1}{m}\sum\limits_{j=1}^{m}p_{j1}, \frac{1}{m}\sum\limits_{j=1}^{m}p_{j2}, \cdots, \frac{1}{m}\sum\limits_{j=1}^{m}p_{jl}).
\end{equation}

Then relationship vector for $n$th label combination $v_n = \{p_{n1}^{v}, p_{n2}^{v}, \cdots, p_{nl}^{v}\}$ can be calculated via normalization operation:

\begin{equation}
p_{ni}^{v} = \frac{p^{avg}_{n}(i)}{\sum_{j=1}^{l}{p^{avg}_{n}(j)}}, i=\{ 1, 2, \cdots, l\},
\end{equation}
where $p_{i}^{v}$ is the $i$th value of $v_n$ and $p^{avg}_{n}(i)$ is the $i$th value of $p^{avg}_{n}$. Vector $v_n$ reflects the probability distribution for $n$th label combination. Vectors set $V =\{v_j|j=1,2,\cdots,N\}$ is defined as the correlation table.

Given a prediction probability vector $\hat{p}$, a normalization operation is also executed to transform $\hat{p}$ into the form of relationship vector as $\hat{p}^{r}$. Then we can search its the top nearest relationship vector in the correlation table, which is obtained through:

\begin{equation}
{\rm K_{(\hat{p}^r,V)}}=\mathop{\arg\min}_{v_{j \in (1,2,\cdots,N)}} L(\hat{p}^{r}, v_{j}),
\end{equation}
where $L$ is the Manhattan distance, and $K$ is the relationship vector $v_n$ that has the minimum distance with $\hat{p}^r$. Then the label combination corresponding to $K_{(\hat{p}^r,V)}$ is the most confident label combination for $\hat{p}^r$.

To obtain a more confident correlation table, we update the table after each AL iteration with a constraint. For relationship vector $v_n^{new}$ in the new table, it replaces the corresponding $v_n$ in the previous table only if it satisfies the following condition:
\vspace{-0.1em}
\begin{equation}
v_n^{new} \cdot z_{n} \geq v_n \cdot z_{n}, \  n = \{1, 2, \dots, N\},
\end{equation}
where $z_{n}$ is the $n$th label combination (\eg, (0, 0, 0, 1)).

\subsubsection{Human-machine interaction}
Given the selected samples and pseudo labels, a HMI is designed for annotators to judge the correctness of annotations and revise the annotations if they are not consistent with the symptom examples. To better understand the HMI, we illustrate the interface in Fig. \ref{interface}. The first row is the pipeline of pseudo-label generation, which is the backend of the interface. The second row is the user interface, which exhibits examples for each label and the selected image. Besides, the annotations would be made as defaults according to the pseudo label. The human annotator only needs to judge whether the default annotations are consistent with the symptom examples.

\subsubsection{CNN network updating}
MSML is updated with a fine-tuning process. At each AL iteration $t$, the CNN is fine-tuned via selected samples. During fine-tuning, only weights of the last three layers in MSML are updated, while the remained weights are frozen to the values from the pre-training. When more images are selected and annotated, the model becomes more robust. The renewed network is exploited to update the state.

\section{Experiments}

\subsection{Implementation Details}
We build the first version of the COVID-19 US dataset, called COVID19-LUSMS v1. US videos are collected in the Third People’s Hospital of Shenzhen, China. A total of $71$ COVID-19 patients are inspected, including $678$ videos. Random rotation (up to 10 degrees) and horizontal flips are used as data augmentation transformations. The Stochastic Gradient Descent optimization is adopted. Learning rate is $2 \times 10^{-3}$, batch size is $32$ and momentum is set as $0.9$. $K_{max}$ is $100$ and the AL iteration limit is set as $20$. For effective comparison, random strategy is implemented to randomly select samples for annotations.

\subsection{Quantitative Analysis }
\subsubsection{MSML}

In Tab. \ref{table:com}, we illustrate the performance comparisons on four baselines including POCOVID-Net\cite{soldati2020proposal}, NNBD \cite{8805336}, VGG16 \cite{Simonyan15},  and ResNet \cite{he2016deep}.
We have the following observations and discussion. (1) Accuracy: the proposed MSML model achieves 100\%, 95.72\%, 80.98\%, and 90.09\% accuracy for A-line, B-line, pleural lesion and pleural effusion, respectively. It shows that the MSML model almost outperforms all baseline models concerning accuracy. 
(2) Sensitivity: MSML model achieves 100\%, 98.78\%, 81.38\%, and 6.08\% sensitivity for A-line, B-line, pleural lesion and pleural effusion, respectively. It performs similar sensitivity with baseline models, which is mainly because of the distinct patterns of these symptoms. Besides, all these methods perform poor sensitivity for pleural effusion. We explain that multiple symptoms appearing simultaneously may cause complicated patterns, especially for pleural effusion.
(3) Specificity: MSML model achieves 100\%, 81.81\%, 80.67\%, and 100\% specificity for A-line, B-line, pleural lesion and pleural effusion, respectively. It shows superior results on all symptoms compared with baseline models, which means it learns a better feature representation compared with baseline models.



\begin{figure}[!t]
	\centering
	\subfigure[Accuracy for A-line]{
		\begin{minipage}[t]{0.45\linewidth}
			\centering
			\includegraphics[width=1\columnwidth]{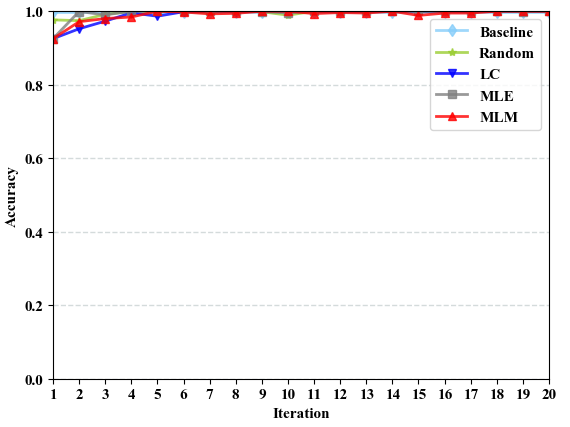}
		\end{minipage}%
	}%
	\subfigure[Accuracy for B-line]{
		\begin{minipage}[t]{0.45\linewidth}
			\centering
			\includegraphics[width=1\columnwidth]{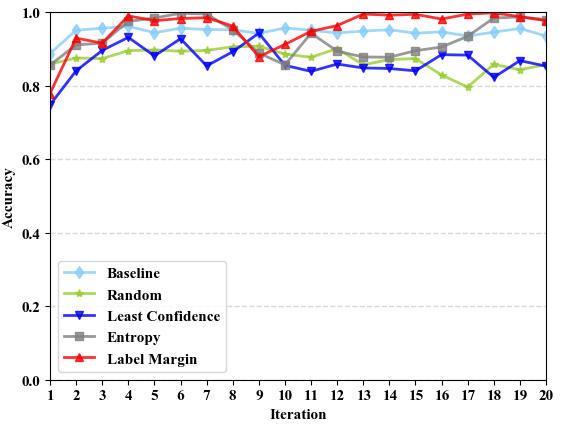}
		\end{minipage}%
	}%
	
	\subfigure[Accuracy for P-effusion]{
		\begin{minipage}[t]{0.45\linewidth}
			\centering
			\includegraphics[width=1\columnwidth]{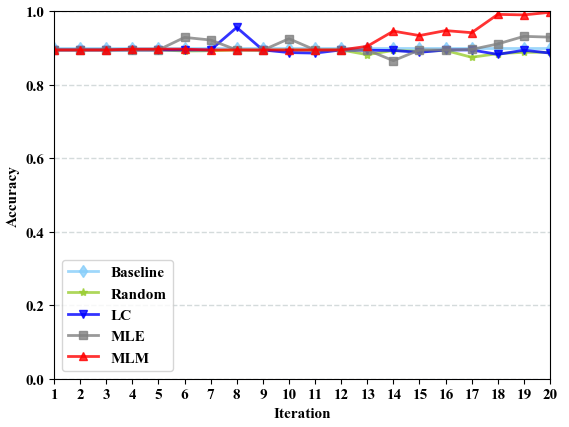}
		\end{minipage}%
	}%
    \subfigure[Accuracy for P-lesion]{
    	\begin{minipage}[t]{0.45\linewidth}
    		\centering
    		\includegraphics[width=1\columnwidth]{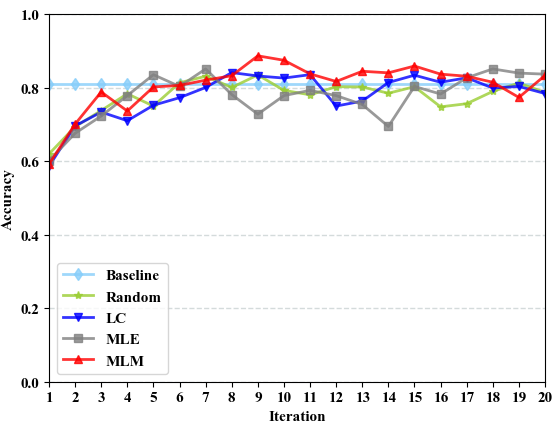}
    	\end{minipage}%
    }%
	\centering
	\caption{Comparisons of the proposed MSML-TSAL method with different selection strategies on accuracy. Baseline means using all labeled images in the dataset without using any selection strategy. Abscissa indicates the training iterations.}
	\label{improve}
	\vspace{-1.0em}
\end{figure}

\begin{figure}[!t]
    \setlength{\abovecaptionskip}{0.cm}
	\centering
	\subfigure[Unselected samples for TSAL]{
		\begin{minipage}[t]{0.5\linewidth}
			\centering
			\includegraphics[width=1.0\columnwidth]{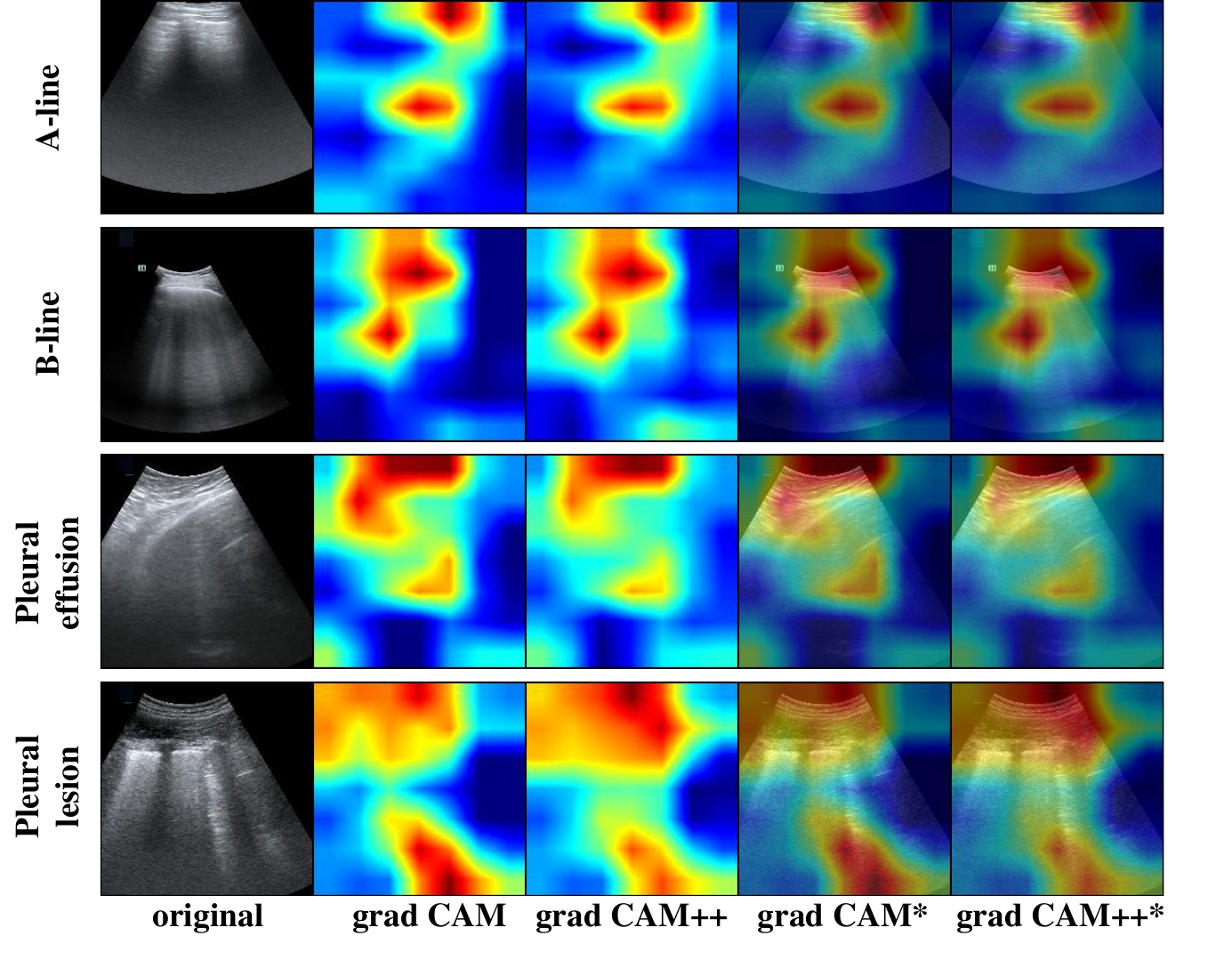}
		\end{minipage}%
	}%
	\subfigure[Selected samples for TSAL]{
		\begin{minipage}[t]{0.5\linewidth}
			\centering
			\includegraphics[width=1.0\columnwidth]{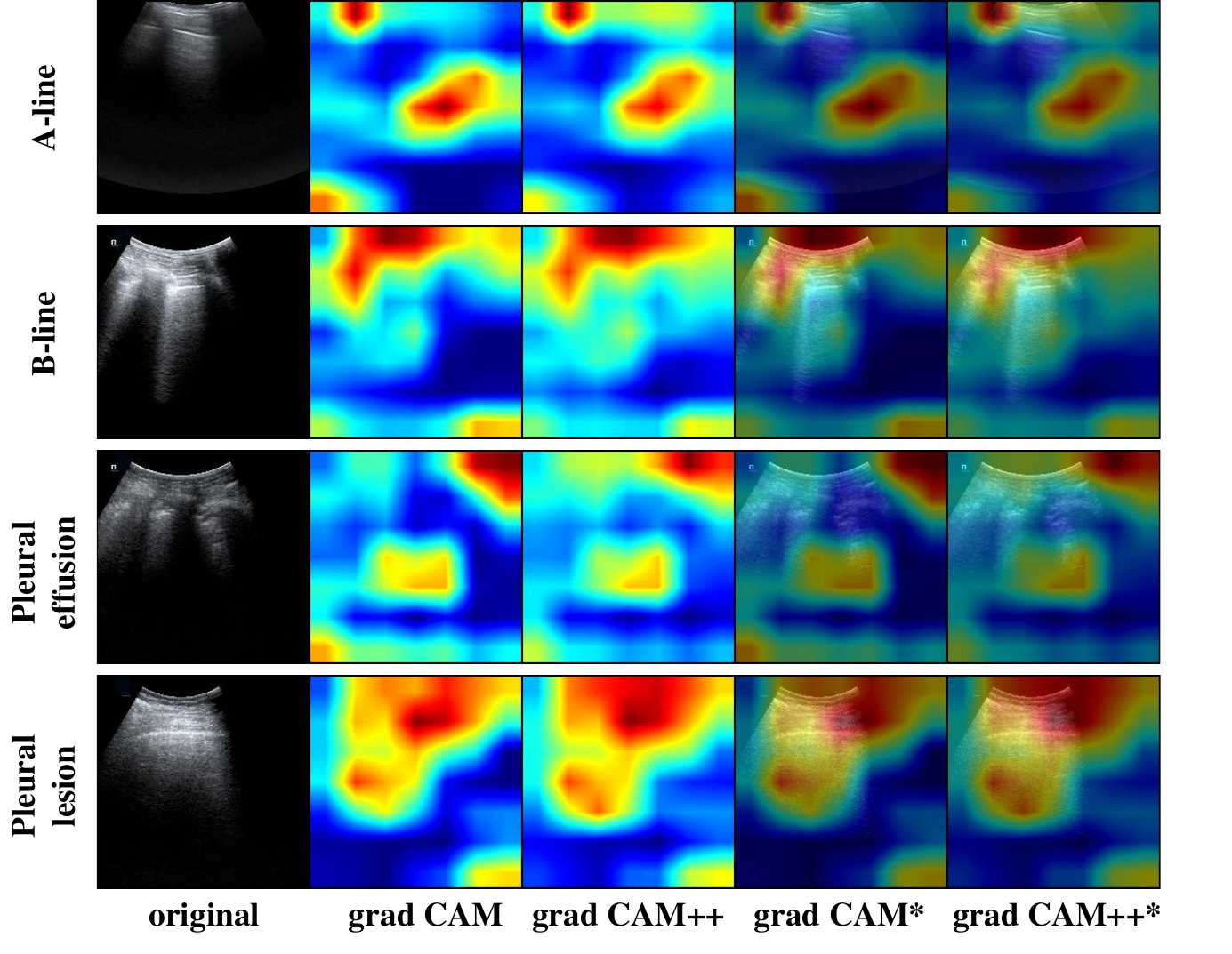}
		\end{minipage}%
	}%
	\centering
	\caption{Visualized results for MSML-TSAL. (a) The training weights of MSML-TSAL is used for the visualization of the unselected samples. (b) The training weights of MSML-TSAL is used for the visualization of the selected sample. (*) denotes that the attention map is overlapped on the original image.}
	\label{vis}
	\vspace{-1.5em}
\end{figure}

\subsubsection{MSML-TSAL} We report the performance of MSML-TSAL concerning four selection strategies in Tab. \ref{table:com}. (1) Accuracy: MLE only uses $27.6\%$ data to train a MSML model, whose accuracy outperforms the baseline models except for pleural effusion. MLM uses $14.7\%$ labeled data to obtain a similar performance as the full training data. Only using $16.6\%$ data, LC obtains comparable accuracy results as the full training set. (2) Sensitivity: all strategies achieve similar sensitivity using fewer images for A-line and B-line, but perform worse for pleural lesion, because the pleural lesion often appears together with other symptoms. It should be mentioned that, for pleural effusion, all strategies obtain near $0$ sensitivity when using less than 30\% data, because the image of pleural effusion is far less than others. (3) Specificity: these strategies merely exploit 16.6\%, 27.6\%, 16.6\%, and 14.7\% data to achieve similar or better specificity performance, among which the random strategy performs worst. Surprisingly, MSML-TSAL improves specificity for B-line and pleural effusion by a large margin, \eg, the specificity for B-line is increased from 81.81\% to 92.49\%. We deduce that the selection strategies can alleviate the unbalanced problem of data distribution.

In Fig. \ref{improve}, we present the accuracy of MSML-TSAL combined with four selection strategies that achieve the best performance on the COVID19-LUSMS dataset. Almost all curves of different strategies achieve similar or better performance compared with the baseline, which uses the full training set (light blue curves). During the whole training process, these curves have several oscillations. We deduce that features in the COVID19-LUSMS dataset are complex and the amount of labeled samples in each iteration is limited. Besides, we note that MLM (red curves) in Fig. \ref{improve} performs the best among the four strategies, with its highest accuracy at the final AL iteration (\ie, the 20th iteration). Concerning smoothness, we can see that the MLM performs the most stable changing tendency.

\subsection{Qualitative Analysis}

We exploit Grad-CAM \cite{Selvaraju_2017_ICCV} to highlight the attention regions in the images. As shown in Fig. \ref{vis}(a), attention regions from the MSML-TSAL model are consistent with the regions from the doctor. For example, based on the prior knowledge of doctors, A-line has an obvious horizontal-line region in the upper part of the images, while the lower part of the image is dark. Corresponding to the attention heatmaps, the visualized attention regions perform consistent results with doctors' 
diagnosis. From Fig. \ref{vis}(b), we find that these images contain large dark regions, which are not pathological changing regions and may cause complicated characteristics for the model learning. These characteristics are not well learned via previous training, because the attention regions are likely to focus on the dark regions as shown in Fig. \ref{vis}(b). Thus, these images should join in further training for their complicated information.


\begin{table}[!t]
	\renewcommand\arraystretch{1}
	\fontsize{10}{11}\selectfont
	\footnotesize
	\centering
	\caption{Amount of manually labeled data.}
	\begin{tabular}{ccccccc}
		\toprule
		Iteration & 1st & 2nd & 3rd & 4th & 5th  \\ \hline
		HMI & 100\% & 100\% & 100\% & 100\% & 100\% \\
		HMI + pseudo-label & 100\% & 29\% & 11\% & 5\% & 1\% \\
		HMI + pseudo-label + CV & 100\% & 2\% & 1\% & 0\% & 0\% \\
		\hline
	\end{tabular}
	\label{Validation}
	\vspace{-1.0em}
\end{table}



\subsection{Component analysis for label stream}
An ablation study is carried out to justify the efficiency of the label stream for pseudo-label assignment.
As shown in Tab. \ref{Validation}, human annotators need to manually annotate all selected images in each iteration, using only HMI without pseudo-label and confidence validation. By pseudo-label, the manual annotations gradually decrease with the improving performance of MSML. Confidence validation can further reduce the manual annotations, \ie, nearly zero after the first iteration. Through label stream, the selected samples can be automatically annotated, thus human annotators only need to confirm in HMI rather than manual annotations.




\section{Conclusion}

To achieve accurate classification of COVID-19 multiple symptoms of the lung US image with less annotated data, we innovatively propose a TSAL framework to effectively train the MSML model with less labeling efforts in a semi-supervised manner. Specifically, we design a MLM strategy and a confidence validation for TSAL by label correlations information. Moreover, a new large-scale lung US image dataset with multiple COVID-19 symptoms is built in this work. Quantitative and qualitative experimental results show that the TSAL model can achieve competitive performance, and we can train an effective MSML model merely using less than 20\% data of the full training set. In future work, it is worthwhile to explore the reinforcement learning to learn a powerful and adaptive policy for image selection.


\bibliographystyle{IEEEtran}
\bibliography{refs}

\end{document}